\documentclass[conference]{IEEEtran}
\IEEEoverridecommandlockouts
\usepackage{cite}
\usepackage{amsmath,amssymb,amsfonts}
\usepackage{algorithmic}
\usepackage{graphicx}
\usepackage{textcomp}
\usepackage{xcolor}
\usepackage{color, colortbl}
\usepackage{url}
\usepackage[caption=false]{subfig}
\usepackage{balance}  
\usepackage[ruled,vlined,linesnumbered]{algorithm2e}
\usepackage{multirow}
\usepackage{hhline}
\usepackage[]{caption}
\def\BibTeX{{\rm B\kern-.05em{\sc i\kern-.025em b}\kern-.08em
    T\kern-.1667em\lower.7ex\hbox{E}\kern-.125emX}}

\definecolor{Gray}{gray}{0.9}
\usepackage{comment}
\usepackage{array, boldline, makecell, booktabs}
\usepackage{adjustbox}
\makeatletter
\newcommand{\removelatexerror}{\let\@latex@error\@gobble}
\makeatother
\makeatletter
\newcommand{\figcaption}{\def\@captype{figure}\caption}
\newcommand{\tabcaption}{\def\@captype{table}\caption}
\makeatother

\begin{document}

\title{Interpreting County Level COVID-19 Infection and Feature Sensitivity using Deep
Learning Time Series Models}


\author{
    \IEEEauthorblockN{
    Md Khairul Islam \textsuperscript{\rm 1},
    Di Zhu \textsuperscript{\rm 1},
    Yingzheng Liu \textsuperscript{\rm 1},
    Andrej Erkelens \textsuperscript{\rm 2} ,
    Nick Daniello \textsuperscript{\rm 2},
    Judy Fox \textsuperscript{\rm 1,2}
    }

    \IEEEauthorblockA{
        \textsuperscript{\rm 1} Computer Science Department, University of Virginia \\
     \textsuperscript{\rm 2} School of Data Science, University of Virginia
      \\
      Charlottesville, USA \\
    Email : \{mi3se, yqx8es, yl4dt, wsw3fa, njd9e, cwk9mp\}@virginia.edu 
    }
}

\maketitle
\begin{abstract}
Interpretable machine learning plays a key role in healthcare because it is challenging in understanding feature importance in deep learning model predictions. We propose a novel framework that uses deep learning to study feature sensitivity for model predictions. This work combines sensitivity analysis with heterogeneous time-series deep learning model prediction, which corresponds to the interpretations of Spatio-temporal features from what the model has actually learned. We forecast county-level COVID-19 infection using the Temporal Fusion Transformer (TFT). We then use the sensitivity analysis extending Morris Method to see how sensitive the outputs are with respect to perturbation to our static and dynamic input features. The significance of the work is grounded in a real-world COVID-19 infection prediction with highly non-stationary, finely granular, and heterogeneous data. 1) Our model can capture the detailed daily changes of temporal and spatial model behaviors and achieves high prediction performance compared to a PyTorch baseline. 2) By analyzing the Morris sensitivity indices and attention patterns, we decipher the meaning of feature importance with observational population and dynamic model changes. 3) We have collected 2.5 years of socioeconomic and health features over 3142 US counties, such as observed cases and deaths, and a number of static (age distribution, health disparity, and industry) and dynamic features (vaccination, disease spread, transmissible cases, and social distancing). Using the proposed framework, we conduct extensive experiments and show our model can learn complex interactions and perform predictions for daily infection at the county level. Being able to model the disease infection with a hybrid prediction and description accuracy measurement with Morris index at the county level is a central idea that sheds light on individual feature interpretation via sensitivity analysis. 
\end{abstract} 

\hfill \break

\section{Introduction}

Interpretation of machine learning models has recently  \cite{AAAI22_Wiegand} led to numerous research applications of AI for social impact. This includes direct analysis of model components with casual inference and uncertainty estimation or studying sensitivity to input perturbations. Typically a simpler model is easier to interpret but can result in lower predictive accuracy. One natural question that arises is how to interpret these complex deep learning models, which may describe the data better. One major challenge of interpretability is the gap between model prediction accuracy and descriptive accuracy in real-world problems. The latter can be illustrated by a quantifiable measurement and explanation of the individual feature importance with regard to the model's forecast relevancy.

To our knowledge, however, no prior studies have evaluated individual feature importance at the county level using deep learning and the Morris method. We have been closely monitoring the scientific literature and identifying reports describing the community-level impact of COVID-19. A number of factors contribute to COVID-19 cases and deaths, including a very diverse set of socioeconomic and geographic-specific features. A more granular real-time analysis that considers important county-level factors is lacking and urgently needed. Furthermore, non-stationary time series (with their distribution drifting over time) \cite{arik2022self} or time series with extreme events  \cite{Peng2021} or unknown events like COVID variants are particularly challenging to model and interpret. 

To effectively study county-level input features, we design a novel method to compute the Morris index but generalize it to multidimensional spatial and temporal variables. Using a self-attention-based Temporal Fusion Transformer (TFT) model \cite{lim2021temporal}, we can capture a complex mix and full range of static and dynamic covariates, known inputs, and other exogenous time series parameters. We perform individual feature importance evaluations to identify the most influential features for prediction and the sensitivity of infected cases. The results show that the model obtains significant performance and learns temporal patterns. More significantly, our scaled Morris index provides sensitivity measurement to individual features that help policymakers develop effective control strategies in response to the rapidly evolving pandemic. We have made our code available on GitHub \footnote{\url{https://github.com/Data-ScienceHub/gpce-sensitivity}}. In summary, we've made the following contributions:

\begin{itemize}
    \item Introduce individual feature sensitivity to forecasting outputs with an extended Morris Method for multidimensional spatial and temporal data.
    \item Model heterogeneous time-series prediction and analyze attention weights for insights on feature importance. 
    \item Stratify county-level population characteristics(Age and Industry segments) from socioeconomic and health data.
\end{itemize}

The rest of the paper is structured as follows. Section \ref{sec:data} discusses the data collection, and feature descriptions. Section \ref{sec:background} presents the background on the TFT model architecture and the Morris method. Section \ref{sec:exp_setup} describes the data pre-processing and experimental setups. Section \ref{sec:attention} analyzes the temporal patterns and feature importance insight from the TFT. Section \ref{sec:feature_sensitivity} discusses the sensitivity analysis with the Morris method. Then Section \ref{sec:related_works} discusses the related works and Section \ref{sec:conclusion} has the concluding remarks and impact on possible future works.

\section{Input Data and Features \label{sec:data}}

\begin{figure}[!ht]
    \centering
    \includegraphics[width=0.47\textwidth]{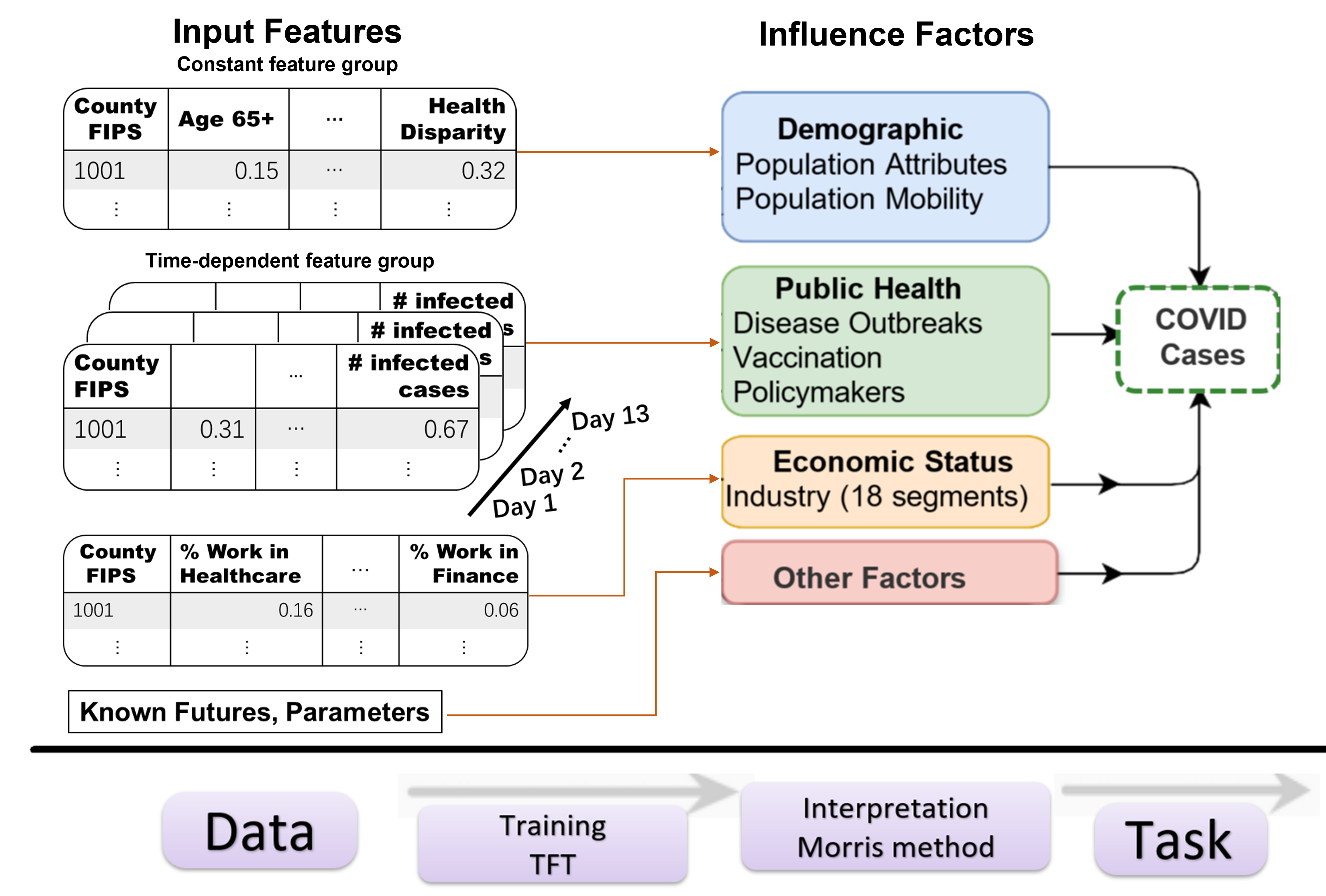}
    \caption{The data: the feature groups and influencing factors.}    
    \label{fig:covid_factors}
\end{figure}

We collected our dataset for 3142 US counties. They are from multiple sources, including CDC (Centers for Disease Control and Prevention), USA Facts \cite{usafacts}, Unacast \cite{unacast}. The dynamic features include entries from 02-29-2020  to 05-17-2022. Except for vaccination, where the earliest available data in CDC \cite{vaccination} was from 12-14-2020 when the US initiated a nationwide COVID-19 vaccination campaign. In total we select 9 observed features, static and dynamic, to predict cases and deaths. Fig.\ref{fig:covid_factors} summarizes the feature groups with the influencing factors they capture and which county characteristics they represent.

Table \ref{table:features} lists the features used by our model and the respective sources with descriptions. In particular, age distribution, health disparities, disease spread, social distancing, and transmissible cases features are collected from the outputs of the COVID-19 Pandemic Vulnerability Index (PVI) dashboard \cite{Marvel2020}, maintained by the National Institute of Health (NIH). 

\begin{table*}[!ht]
\centering
\caption{Short description of input feature groups (and target features). Refer to references and Appendix for full details. }

\begin{tabular}{p{2.4cm}| p{9cm} |p{2.1cm} |p{2cm}}
\Xhline{1pt}
\textbf{Feature} & \textbf{Description} & \textbf{Data Source} & \textbf{Input Type} \\ \hline
Cases &  Daily COVID-19 cases  & \multirow{2}{*}{USA Facts \cite{usafacts}} & \multirow{2}{*}{Target} \\
Deaths & Daily COVID-19 deaths  & & \\ \hline
Age Distribution & Percentage of population aged 65 or older & \multirow{2}{*}{SVI \cite{svi2018}} & \multirow{3}{*}{Static} \\

Health Disparities &  Uninsured population percent and socioeconomic status & & \\
Industry & Percentage of population in different industry sectors (only used in Section \ref{sec:feature_sensitivity}) & Census Bureau \cite{industry2017} & \\ \hline
Vaccination & Percentage of population fully vaccinated  & CDC \cite{vaccination} & \\ 
Disease Spread &  Fraction of total cases from the last 14 days (one incubation period) & USA Facts \cite{usafacts} & \multirow{3}{*}{Observed} \\ 
Transmissible Cases &  Population size divided by cases from the last 14 days & USA Facts \cite{usafacts} & \\
Social Distancing  &  Change in distance travelled relative to baseline(previous year), based on cell phone mobility data  &  Unacast \cite{unacast} & \\ \hline
SinWeekly &  $\sin$ (day of the week/7) & Date & \multirow{3}{*}{Known Future} \\ 
CosWeekly &  $\cos$  (day of the week/7) & Date &  \\
Linear Space & Unique index for each county. & USA Facts \cite{usafacts} & \\ 
\Xhline{1pt} 
\end{tabular}
\label{table:features}
\end{table*}

\section{Background and Theoretical Foundation \label{sec:background}}
\subsection{Temporal Fusion Transformer}
\begin{figure}[!h]
    \centering
    \includegraphics[width=0.45\textwidth]{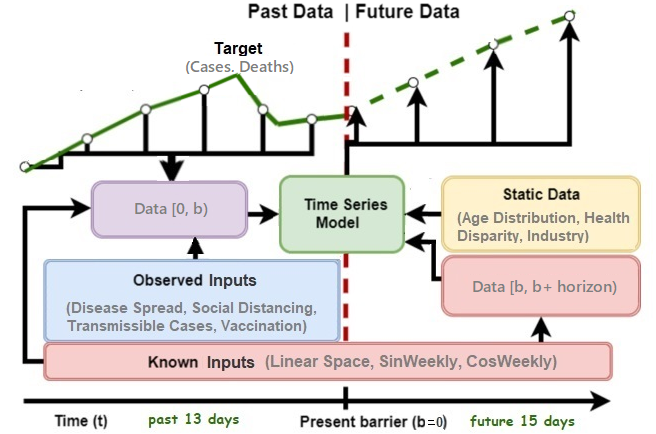}
    \caption{A time series forecasting model. Each sliding window consists of time-sequential data that is split into two parts, the past, and the future.}
    \label{fig:TFT_data_model}
\end{figure}

We used the TFT model \cite{lim2021temporal} to predict daily COVID-19 cases and deaths at the county level.
For this work, we dive deeper into the COVID-19 daily cases prediction and combine sensitivity analysis of individual features. Figure  \ref{fig:TFT_data_model} shows a high-level overview of the work. Gated Residual Network (GRN) is the building block of TFT and it enables more efficient use of the model architecture. TFT takes static metadata, time-varying past inputs, and time-varying known future inputs. The model inputs are passed through a Variable Selection Network (VSN) to select the most salient features and filter out noise. 

Learning significant data points is done by leveraging local context with LSTM-based sequence to sequence layer. Past inputs are fed into the encoder, whereas known future inputs are fed into the decoder. Their outputs go through a static enrichment layer which enhances temporal features with static metadata. Following static enrichment, TFT adds a novel interpretable multi-head self-attention mechanism to better learn the different temporal patterns. This allows TFT to learn long-rage dependencies that can be challenging for Recurrent Neural Network (RNN) based models. Following the self-attention layer, additional gating layers are added to facilitate training.

\subsection{Sensitivity Analysis and The Morris Method}
\label{subsec:Morris}
Sensitivity Analysis is the study of the input-output relationship in a computational model \cite{saltelli2008global}. It can identify the importance of each model parameter in determining the outputs. \cite{brunner2019identifiability} proposed gradient-based attribution which approximates the neural network function $f(\boldsymbol{X})$ around a given input $\boldsymbol{X}$ by linear part of the Taylor expansion as
\begin{equation}
    \label{eqn:taylor}
    f(\boldsymbol{X}+\Delta \boldsymbol{X})\approx f(\boldsymbol{X})+\nabla_{\boldsymbol{X}}f(\boldsymbol{X})^{T}\cdot \Delta \boldsymbol{X}
\end{equation}
and they analyzed the network sensitivity by looking at how small changes $\Delta \boldsymbol{X}$ at the input correlate with changes at the output. Gradient $\nabla_{x_i}f=\frac{\delta f(\boldsymbol{X})}{\delta}$ gives the linear approximation of this change for a change in the $i$-th input token $x_i \in \mathbf{R}$, and the attribution of how much input token $x_i$ affects the network output $f(\boldsymbol{X})$ can be approximated by the L2 norm of $\nabla_{x_i}f$.

\subsection{Problem Statement}
We will use deep learning to study feature sensitivity for model predictions of COVID-19 infection at the county level. Given this, we adopted the Morris method \cite{morris1991factorial}, a reliable and efficient sensitivity analysis method that defines the sensitivity of a model input as the ratio of the change in an output variable to the change in an input feature. More precisely, given a model $\textit{\textbf{Y}}=f(\textit{\textbf{X}})$, the sensitivity (or the elementary effect) of a model input feature $x_{i}$ can be defined as 
\begin{equation}
\label{eqn:morris}
    EE_{i}(\boldsymbol{X})=\frac{y(x_{1}, x_{2},\dots,x_{i}+\Delta,\dots,x_{k}) - y(\boldsymbol{X})}{\Delta}
\end{equation}
where $\boldsymbol{X}$ is a scaled vector of \textit{k} parameters and $\Delta$ is the change to an input feature. Since elementary effects may cancel each other out, the mean of the absolute values in distribution $EE_{i}(\boldsymbol{X})$, denoted by $\boldsymbol{\mu^{*}}$ (called \textbf{the Morris Index}), is recommended because it provides true importance of features \cite{campolongo2007effective}.

\SetKwComment{Comment}{/* }{ */}
\begin{figure}[!h]
\small
    \renewcommand{\algorithmicrequire}{\textbf{Input:}}
    \renewcommand{\algorithmicensure}{\textbf{Output:}}
    \removelatexerror
    \begin{algorithm}[H]
        \label{alg:norm_morris}
        \caption{Novel Morris Index Calculation for \textbf{Spatio-temporal} data}
        \KwIn{$\boldsymbol{X}=\{x_{1}, x_{2},\dots,x_{k}\}$, target feature $x_{i} \in \boldsymbol{X}$ with dimension $[C, T]$, model $y$, $\Delta$}
        \tcp*[h]{$\boldsymbol{X}$ is a set of $k$ input features, $\Delta$ is the change to $x_{i}$} \\        
        {$\boldsymbol{Y}_{\Delta} = y(x_{1}, x_{2},\dots,x_{i}+\Delta,\dots,x_{k})$} \\
        {$\boldsymbol{Y} = y(\boldsymbol{X})$} \\
        \While (/* Temporal */) {$t < T$} {
            \tcp*[h]{Loop through 640 Days} \\
            \While (/* Spatial */) {$c < C$} {
                \tcp*[h]{Loop through 3142 US Counties} \\
                {$G \gets G + \lvert\boldsymbol{Y}_{\Delta}[c][t] -  \boldsymbol{Y}[c][t]\rvert$} /*Total Change*/\\
                {$c \gets c+1 $} \\
            }
            {$t \gets t+1 $} \\
            {$c \gets 0 $} \\
        }
        \tcp*[h]{Calculate normalized Morris Index $\hat{\mu^{*}}$} \\
        {$\hat{\mu^{*}} = G / (C*T*\Delta)$}  \\
        \algorithmicreturn  $\hat{\mu^{*}}$
    \end{algorithm}
\end{figure}
The original Morris method was proposed to screen static input factors (or static features) but this is not conventional for the time series dataset (or dynamic features) where we look at time variation and spatial variation with an important overall influence on the output of COVID-19 cases prediction using TFT. Hence, we design and implement a revised Morris Algorithm \ref{alg:norm_morris} to handle the Spatio-temporal COVID-19 data sequences. The algorithm calculates a \textbf{normalized Morris Index} $\hat{\mu^{*}}$ by dividing the total change to the output $G$ by the total number of counties $C$, the total number of daily timestamps $T$ and the change to the input $\Delta$. In this study, $C$ is the total number of counties and $T$ is the total number of daily timestamps between 2-29-2020 and 11-29-2021. 
\section{Experimental Setup}
\label{sec:exp_setup}

\subsection{Computational Resources}
We implement our TFT model with both Tensorflow \cite{lim2021temporal} and PyTorch \cite{tft_pytorch}. Then we conducted a performance evaluation of the model training on Google Colab and HPC clusters including the GPU nodes in Table \ref{table:environment}. The model training time is about 30 hours. Each training epoch takes on average 50 minutes on a GPU node with at least 32GB of RAM. Each Morris runs with a trained model, and with additional feature analysis that takes around 35 minutes.

\begin{table}[!ht]
\centering
\caption{Runtime environment and hardware specification.}
\small
\begin{tabular}{c|c|c|l}
\hline
\textbf{Driver} & \textbf{CUDA} & \textbf{Processor} & \textbf{NVIDIA GPU} \\ \hline
 \multirow{4}{*}{470.82.01} & \multirow{4}{*}{11.4} & \multirow{4}{*}{Intel Xeon} & A100-SXM4-40GB \\ \cline{4-4}
 & & &Tesla P100-PCIE \\ \cline{4-4}
 && &Tesla V100-SXM2  \\\cline{4-4}
 & & &Tesla K80 \\ \hline
\end{tabular}
\label{table:environment}
\end{table}

\subsection{Evaluation Metrics}
\label{sec:metrics}
Our forecasting models are evaluated using the following metrics. Mean Squared Error (MSE) is used as the loss function following prior works on COVID-19 forecasting \cite{hssayeni2021forecast, arik2022self}. Other metrics include Mean Absolute Error (MAE), Root Mean Square Error (RMSE), Symmetric Mean Absolute Percentage Error (SMAPE), and Normalized Nash-Sutcliffe Efficiency (NNSE) \cite{nossent2012application}.

These metrics have been widely used in evaluating regression model performance \cite{zeroual2020deep}. The benefit of using NNSE is its robustness to error variance. 
NNSE is 1 for a perfect model. A model with an error variance equal to that of the observed time series will give NNSE = 0.5 (NSE=0). When the error variance is larger, NNSE will be in the range (0, 0.5). Also note that MAE favors small population counties more than MSE, where large population counties dominate the error evaluation. SMAPE is very sensitive to the quality of model fitting in small counties. Hence, we can order the metrics (RMSE, NNSE, MAE, and SMAPE) in their evaluation reliability. 

\subsection{Pre-processing}
\begin{figure}[!ht]
\centering
\includegraphics[width=0.4\textwidth]{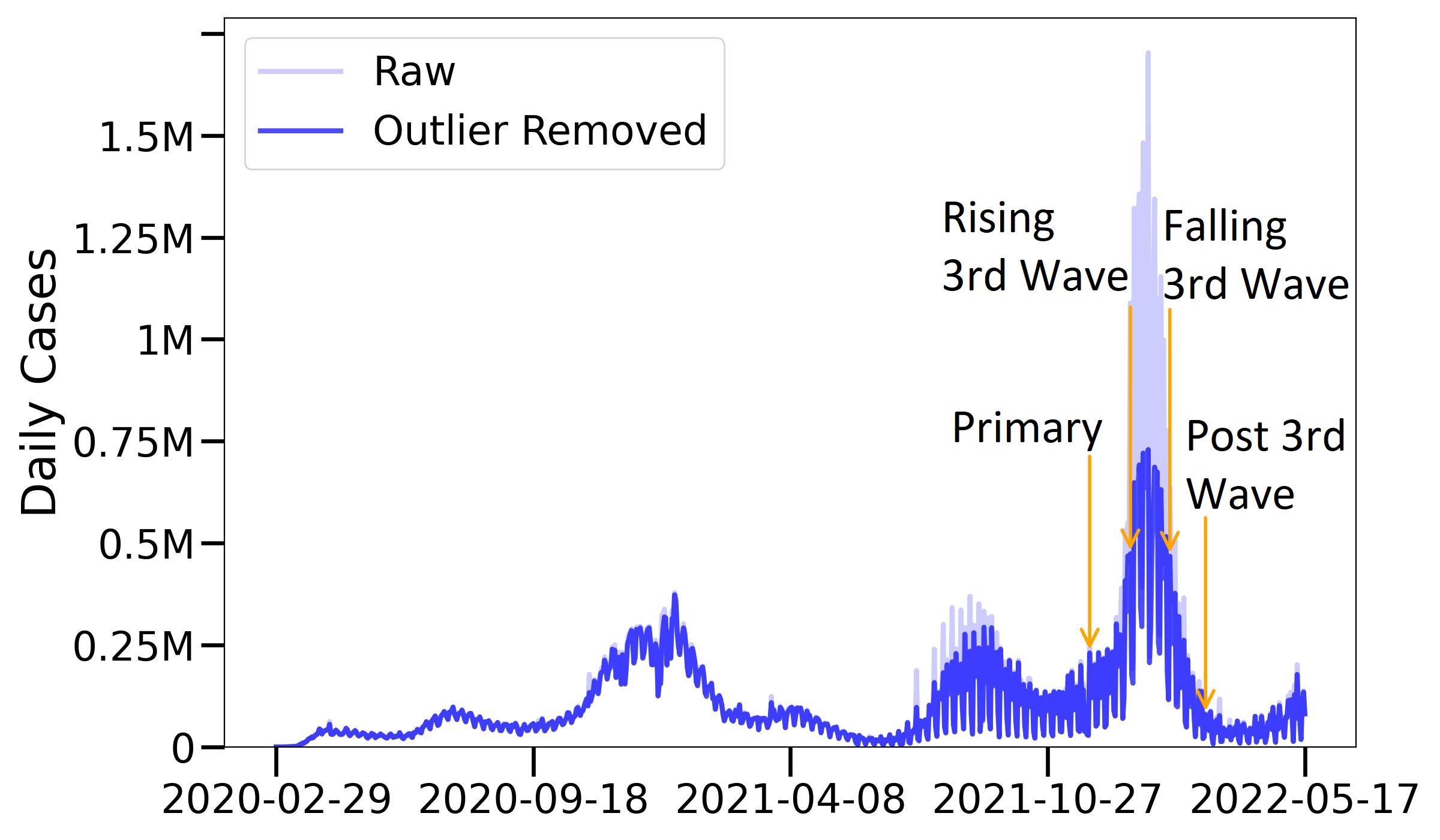}
\caption{Ground truth of reported COVID-19 cases \cite{usafacts} along with dataset splits.}
\label{figure:ground_truth}
\end{figure}

The original data contains outliers caused by rare events or human errors in data collection. We clean the outliers from our input features using the following lower and upper thresholds,

\begin{equation}
\begin{aligned}
\text{lower} = Q1-(7.5*IQR) \\
\text{upper} = Q3+(7.5*IQR)
\end{aligned}
\end{equation}

where $Q1, Q3$ are the first and third percentiles and $IQR$ is the interquartile range. The data statistics before and after removing the outliers are reported in Table \ref{table:statfeatures} and the ground truth is plotted in Figure \ref{figure:ground_truth}. Both input and target features were min-max scaled before feeding to the model. We did not use the moving average to further smooth this dataset because that would filter out persistent temporal patterns observed in the raw data. 

\begin{table}[!ht]
\centering
\caption{Statistics of input features. Cases and vaccination have much higher variance than others.}
\small
\begin{tabular}{p{2.6cm}|cc|cc} 
 \hline
 \multirow{2}{*}{\textbf{Feature}} & \multicolumn{2}{c}{\textbf{Original}} & \multicolumn{2}{c}{\textbf{Cleaned}} \\ \cline{2-5}
 & \textbf{Mean}  & \textbf{Std} & \textbf{Mean}  & \textbf{Std} \\ \hline 
\textbf{Cases} & \textbf{31.67} & \textbf{337.4} & \textbf{27.18} & \textbf{174.2} \\
Deaths & 0.378 & 2.853 & 0.239 & 2.244 \\ \hline
Age Distribution & 0.576 & 0.094 & 0.576 & 0.094 \\
Health Disparities & 0.368 & 0.198 & 0.368 & 0.198 \\  \hline
\textbf{Vaccination}  & \textbf{20.61} & \textbf{22.92} & \textbf{20.61} & \textbf{22.92} \\
Disease Spread & 0.150 & 0.194 & 0.150 & 0.193 \\
Social Distancing & 0.784 & 0.228 & 0.795 & 0.229 \\
Transmissible Cases &  0.492 & 0.210 & 0.491 & 0.210 \\
 \hline
\end{tabular}
\label{table:statfeatures}
\end{table}

\subsection{Benchmark}
We compare our TFT model performance with a simple PyTorch \textit{Baseline} model that predicts future cases and deaths by repeating the last known observations. So there are no parameters to learn for this baseline model. Note that the \textit{std} of "Cases" in our dataset is widely volatile with factors 10X and 5X for raw and outlier removed data respectively. This implies that prediction without 7-day averaging (as in our benchmark) will be 2X to 5X variation from the ground truth, while the smoothed data will be varying by a factor of 2X to 3X at the peak for county-level daily cases prediction.

\paragraph{Train-Validation-Test Split}
We partitioned our dataset into train, validation, and testing sets following the split reported in Table \ref{table:split}. All experiments in this work follow the primary split unless explicitly mentioned. The COVID-19 cases in the US peaked on January 15, 2022, due to new COVID-19 variants \cite{usafacts} like delta and omicron. This period is labeled as the third COVID-19 wave \cite{ELSHABASY2022161} and we named our additional dataset splits based on the part of this third wave included in it. 

\begin{table}[!ht]
\centering
\caption{Dataset splits and dates.}
\small
\begin{tabular}{l|cc|l}
\hline
\textbf{Split} & \textbf{Start} & \textbf{End} & \textbf{Dataset} \\
\hline
\multirow{3}{*}{Primary} &  02-29-2020 & 11-29-2021 & Train \\
& 11-30-2021 & 12-14-2021 & Validation \\
& 12-15-2022 & 12-29-2022 & Test \\ \hline
\multirow{3}{*}{Rising 3rd Wave} &  02-29-2020 & 12-31-2021 & Train \\
& 01-01-2022 & 01-15-2022 & Validation \\
& 01-16-2022 & 01-30-2022 & Test \\ \hline
\multirow{3}{*}{Falling 3rd Wave} &  02-29-2020 & 01-31-2021 & Train \\
& 02-01-2022 & 02-15-2022 & Validation \\
& 02-16-2022 & 03-02-2022 & Test \\ \hline
\multirow{3}{*}{Post 3rd Wave} &  02-29-2020 & 02-28-2021 & Train \\
& 03-01-2022 & 03-15-2022 & Validation \\
& 03-16-2022 & 03-30-2022 & Test \\
\hline
\end{tabular}
\label{table:split}
\end{table}

\paragraph{Model Parameters and Tuning} 
Our TFT models take the prior 13 days of data as input and use that to predict both cases and deaths for the next 15 days. Therefore, at least 13 days of observed data are needed to start giving predictions. Table \ref{table:parameters} presents the parameters for the TFT model and a list of hyper-parameter tuning using the validation set. 

\begin{table}[!ht]
\centering
\caption{TFT model training and network parameters. For hyper-parameter tuning, the list is added and optimal values are in bold.}
\small
\begin{tabular}{l|l|l|l}
\hline
\textbf{Parameters} & \textbf{Value} & \textbf{Parameters} & \textbf{Value}  \\
\hline
learning rate & [\textbf{1e-3}, 1e-4] & batch size & 64 \\
hidden layer size & [\textbf{16}, 32, 64] & dropout rate & 0.2 \\
attention head size & [1, \textbf{4}]  & optimizer & adam \\
gradient clip norm & [0.01, \textbf{1.00}] & loss & MSE \\
\hline
\end{tabular}
\label{table:parameters}
\end{table}

\paragraph{Training and Prediction Performance} 

\begin{figure*}[!ht]
\centering
\subfloat[Train \label{figure:train_performance}]{\includegraphics[width=0.32\textwidth]{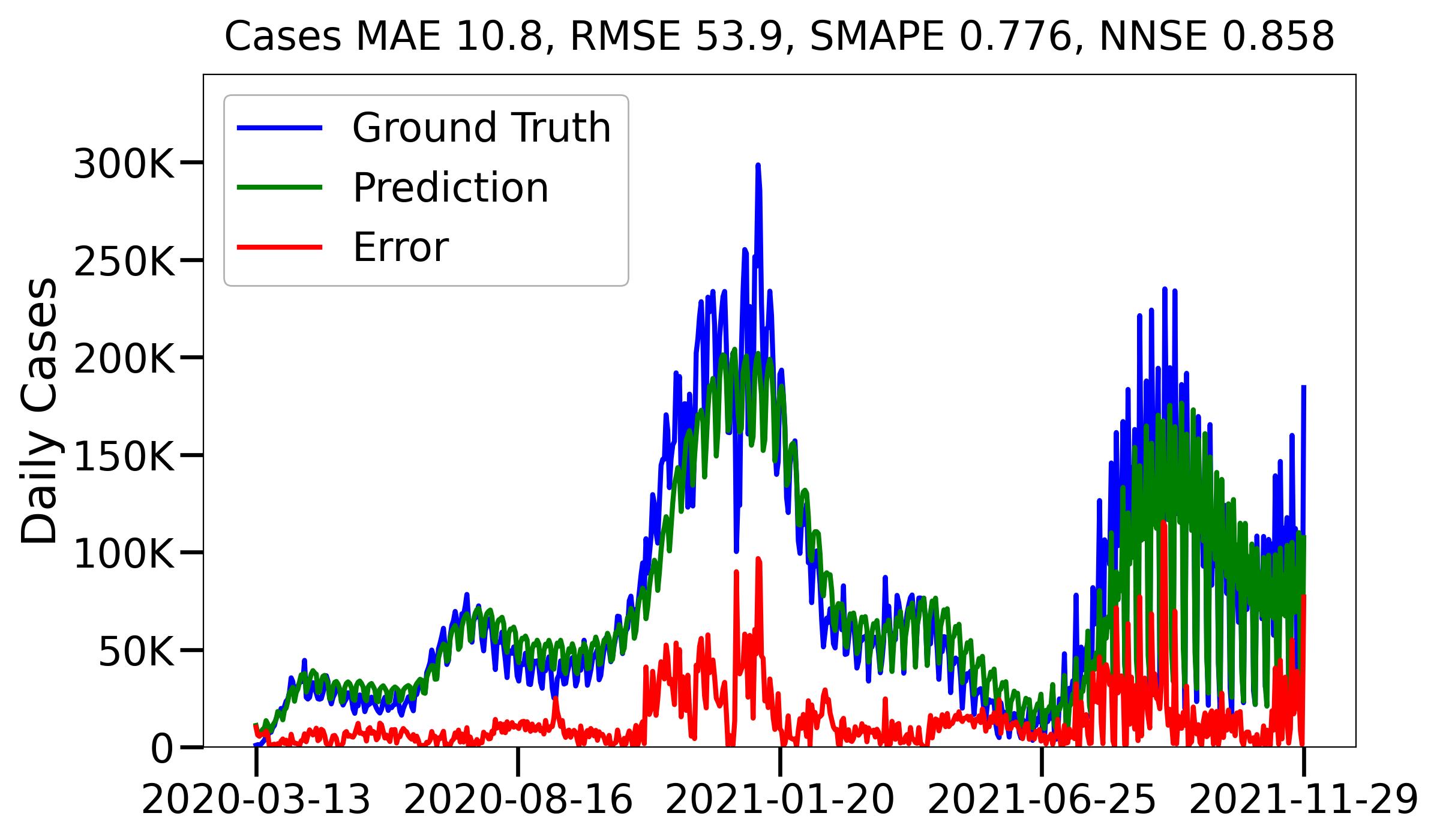}}
\subfloat[Test]{\includegraphics[width=0.60\textwidth]{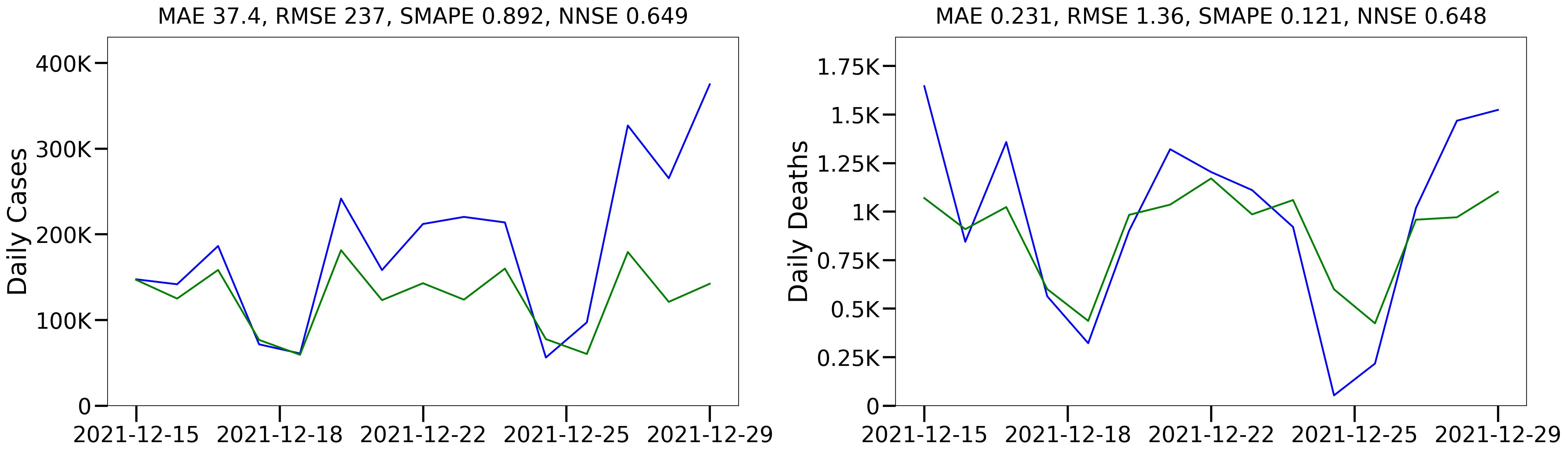}}
\caption{TFT performance based on the primary split (aggregated over 3142 US counties). Case training performance is reported since it is used in the feature sensitivity analysis.}
\label{figure:baseline}
\end{figure*}

Table \ref{table:test_performance} shows that TFT outperforms the baseline model in all metrics. A lower score is better for MAE, RMSE, and SMAPE. Higher is better for NNSE. The losses are calculated using MSE at individual county and date levels. The aggregated prediction plots are presented in Figure \ref{figure:baseline}. 
\begin{table}[!ht]
\centering
\caption{Training and test results and comparison.}
\small
\begin{tabular}{@{}ll|cccc@{}}
\hline
\textbf{Target} & \textbf{Model}  & \textbf{MAE} & \textbf{RMSE} & \textbf{SMAPE} & \textbf{NNSE}\\
\hline
\multirow{2}{*}{Cases} & TFT & 37.40 & 237.0 & 0.892 & 0.649 \\
 & Baseline & 48.47 & 269.0 & 0.921 & 0.588 \\ \hline
\multirow{2}{*}{Deaths} & TFT & 0.231 & 1.36 & 0.121 & 0.648 \\
 & Baseline & 0.346 & 2.25 & 0.139 & 0.401 \\
\hline
\end{tabular}
\label{table:test_performance}
\end{table}


\begin{figure*}[!ht]
    \centering
    \includegraphics[width=0.8\textwidth]{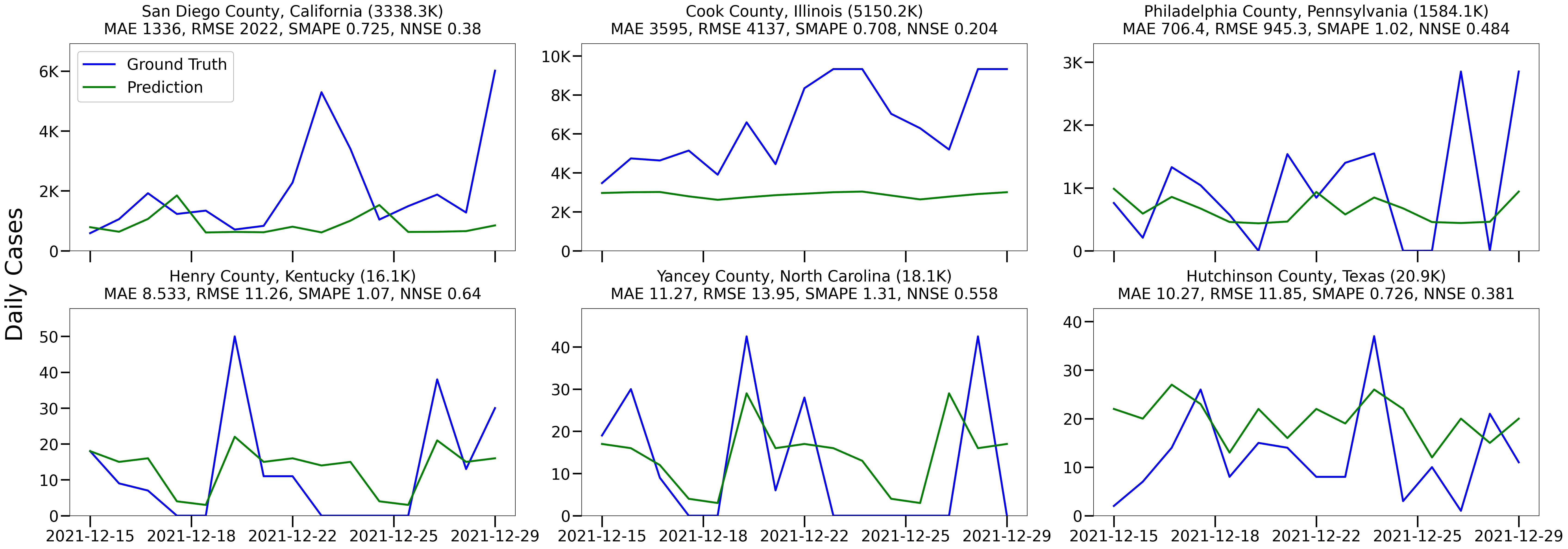}
    \caption{Spatial: Cases prediction performance for six randomly selected US counties. The top row contains counties selected from the Top 100 US counties by population. The bottom row counties are selected from the rest. The county population is reported within brackets.}
    \label{figure:individual_counties}
 \end{figure*}
 
\begin{figure*}[!ht]
 \centering
\includegraphics[width=0.95\textwidth]{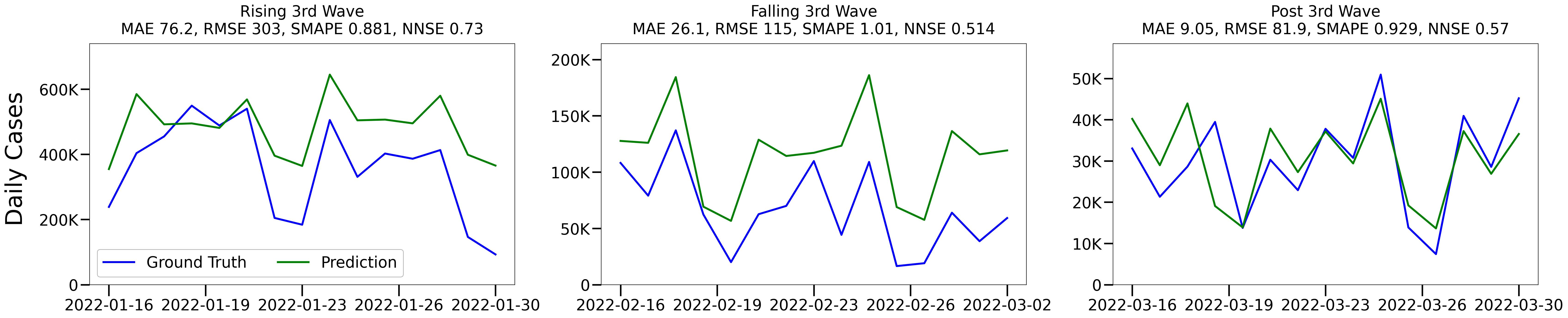}
\caption{Temporal: Cases prediction performance on all US counties for additional test data splits.}
\label{figure:splits}
\end{figure*} 

We further report two experiments: 1) Geo-spatial partition with representative large and small counties, 2) Temporal partition with three additional dataset splits. Figure \ref{figure:individual_counties} shows how the TFT model performs for individual counties with different population sizes. The small counties often have zero COVID-19 cases reported, whereas large counties show sudden large spikes. These features make the prediction more challenging.  

The test results are presented in Figure \ref{figure:splits} for the additional splits, after extending the dataset to more recent dates as reported in Table \ref{table:split}. This shows that the model performs consistently during different phases of COVID-19 waves.  

\section{Attention Weights and Analysis \label{sec:attention}}
Here we present the interpretability use cases of the TFT model in predicting COVID-19 infections. We demonstrate the use cases by (1) analyzing input variable importance, and (2) learning temporal patterns from the dataset. This gives more insight into the significant features and seasonality TFT learns from the dataset.

\subsection{Persistent Temporal Patterns} 

\begin{figure*}[!ht]
 \centering
\includegraphics[width=0.75\textwidth, height=0.30\linewidth]{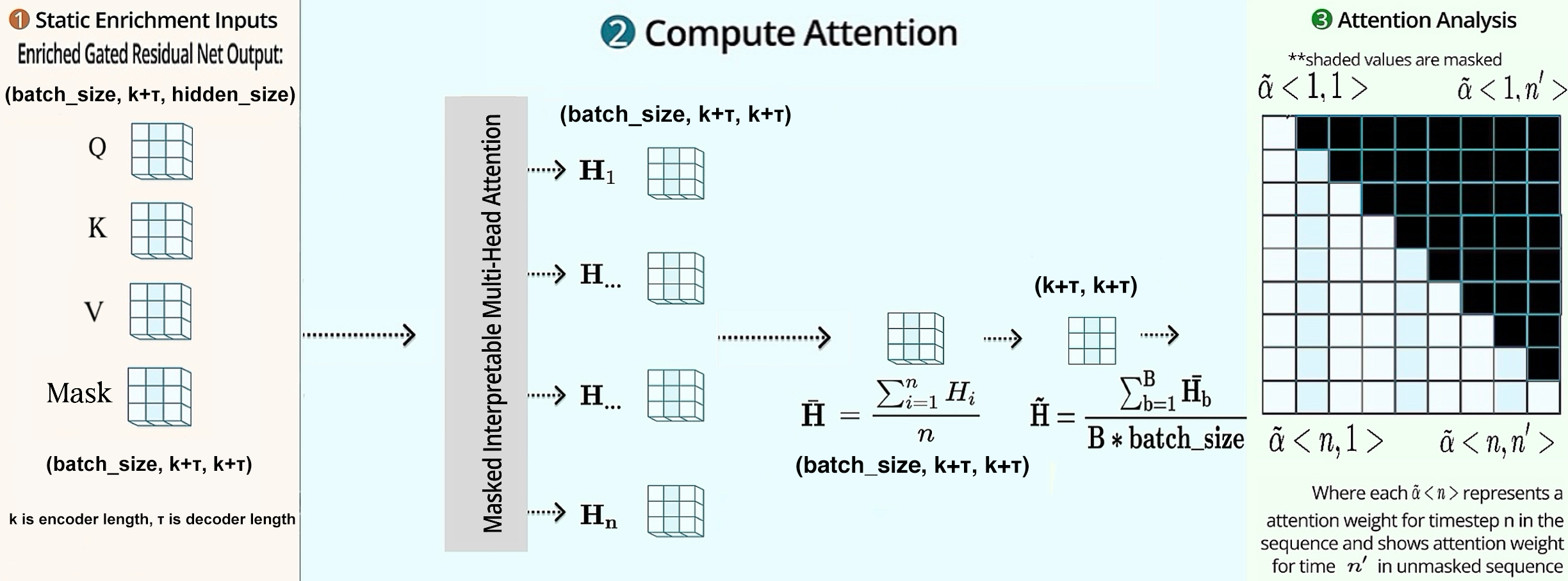}
\caption{Flow of aggregation and selection for TFT attention weights.}
\label{fig:TFTAttention}
\end{figure*}

\begin{figure*}[!ht]
\centering
\subfloat[Attention weights aggregated by past time index. \label{fig:attention_summed} ]{ \includegraphics[width=0.38\linewidth]{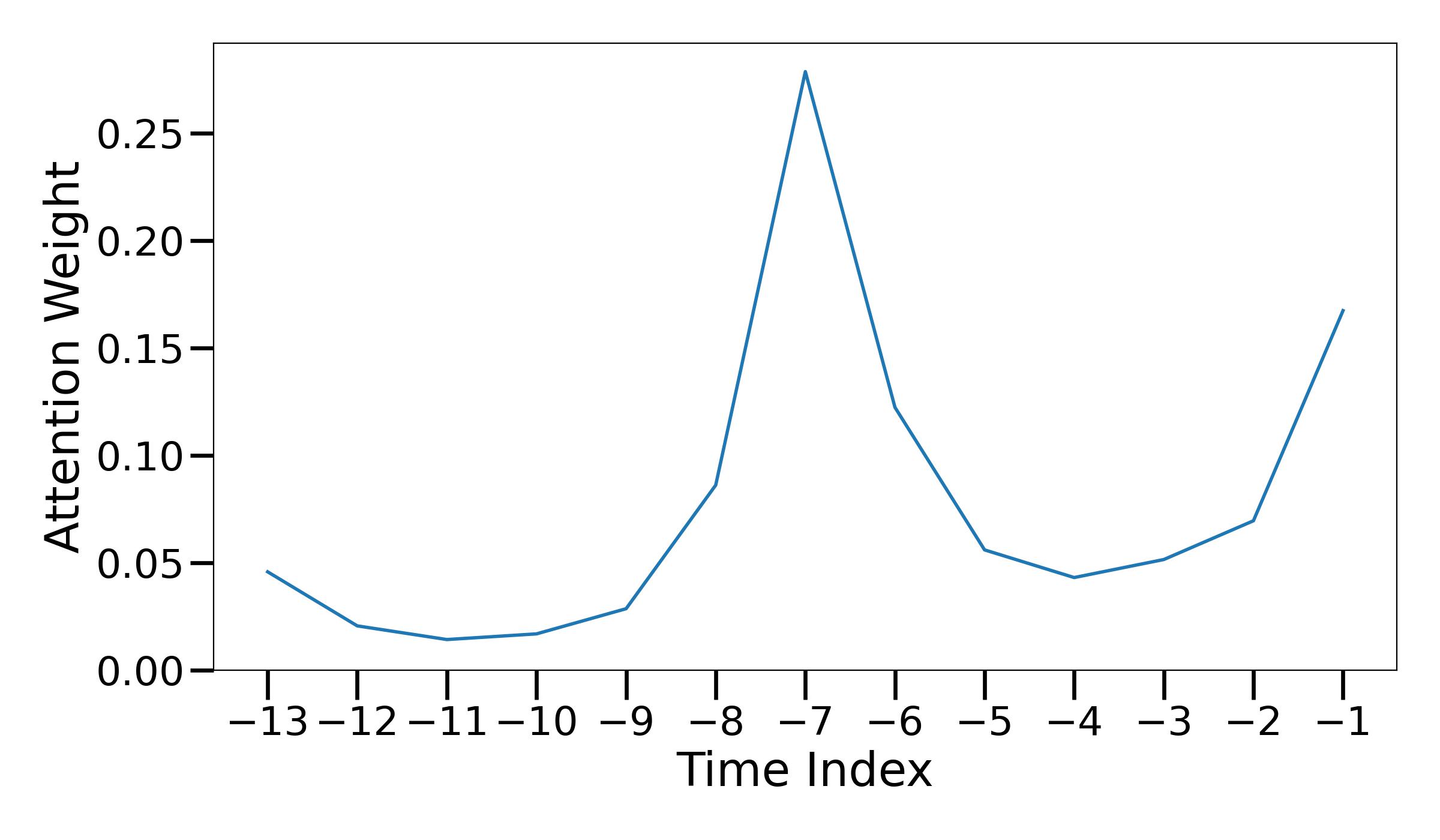}}
\subfloat[Daily cases and attention weights on special events. \label{fig:attention_holiday}]{ \includegraphics[width=0.41\linewidth]{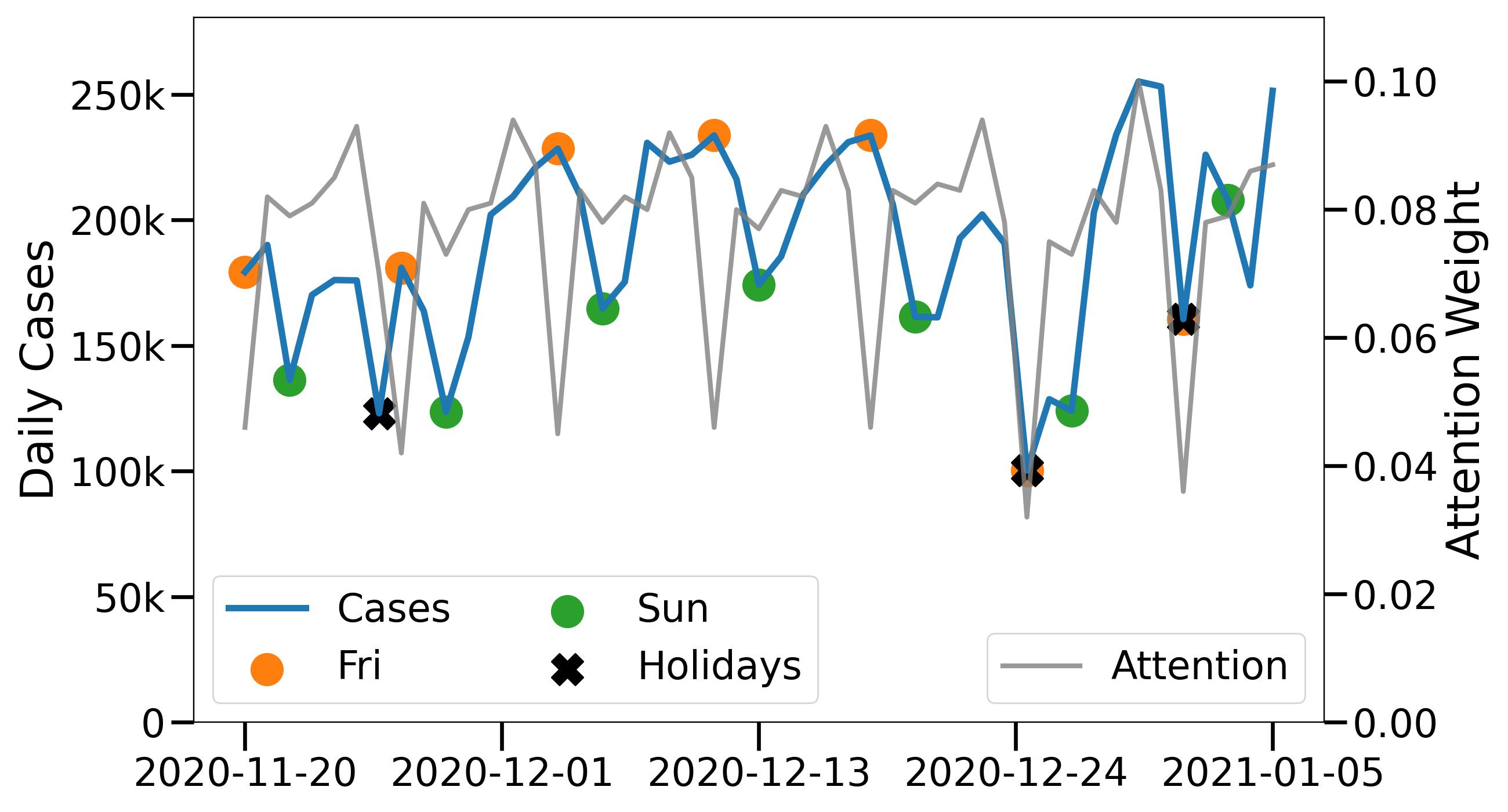}}
\caption{Persistent temporal patterns for one-step-ahead forecast. (a) Report from the same day in the previous week (time index -7) is the most critical in predicting COVID-19 infection. The most recent day in the past is important too, as attention increases close to the present day (time index 0). (b) The attention drops before weekends and holidays when fewer cases are reported.}
\label{fig:attention}
\end{figure*}

Analyzing persistent temporal patterns is a key to understanding the time-dependent relationships present in a given dataset. To improve the interpretability of such patterns the TFT model uses modified multi-head attention that shares values in each head and additively aggregates them:
\begin{equation}
    \text{InterpretableMultiHead}(\boldsymbol{Q},\boldsymbol{K},\boldsymbol{V})=\Tilde{\boldsymbol{H}}\boldsymbol{W}_{H}
\end{equation}
\begin{equation}
    \label{eqn:interpretable}
    \begin{aligned}
    \Tilde{\boldsymbol{H}} &=\Tilde{A}    (\boldsymbol{Q},\boldsymbol{K})\boldsymbol{V}\boldsymbol{W}_{V} \\
    &= \left\{ 1/H\sum^{m_H}_{h=1}A\left(\boldsymbol{Q} \boldsymbol{W}_{Q}^{(h)}, \boldsymbol{K} \boldsymbol{W}_{K}^{(h)}\right) \right\}\boldsymbol{V}\boldsymbol{W}_{V} \\
    &= 1/H\sum^{m_H}_{h=1}\text{Attention}(\boldsymbol{Q}\boldsymbol{W}_{Q}^{(h)},\boldsymbol{K}\boldsymbol{W}_{K}^{(h)}, \boldsymbol{V}\boldsymbol{W}_{V})
    \end{aligned}
\end{equation}

where $\boldsymbol{W}_{Q}^{(h)} \in \mathbf{R}^{d_{model} \times d_{attn}}$, and $\boldsymbol{W}_{K}^{(h)} \in \mathbf{R}^{d_{model} \times d_{attn}}$ are head-specific weights for keys ($K$) and queries ($Q$). A() is a softmax normalization function and $m_H$ is the number of attention heads. $\boldsymbol{W}_{V} \in \mathbf{R}^{d_{model} \times d_{v}}$ is for weights shared across all heads and $\boldsymbol{W}_{H} \in \mathbf{R}^{d_{attn} \times d_{model}}$ is for final linear mapping.

When representing the weights of our self-attention mechanism from equation \ref{eqn:interpretable}, we are left with a $(\pmb{H}, n_{s}, d_{s}, d_{s})$ array, where $\pmb{H}$ is the vector of attention heads, $n_{s}$ is the total number of sequences in our dataset, and $d_{s}$ is the combined input and output sequence length. $n_{s}$ can be calculated by taking $n_{l} * (d - s + 1)$, where $d$ is the number of time steps per ID, and $n_{l}$ is the number of IDs (which is \textbf{counties or FIPS in the COVID-19 prediction}). By taking the mean over $\pmb{H}$ and $n_{s}$ we obtain $\bar{\pmb{H}}$, which leaves us with a high-level view of temporal patterns in each sequence. Left with an $(d_{s}, d_{s})$ array, which is a lower triangular matrix because of the attention mask, we can see the weight at each available day $d_{c}$ attributed to any other visible day $d_{c-i}$ in the sequence. There is an illustration of this in Figure \ref{fig:TFTAttention}.

Figure \ref{fig:attention_summed} presents the mean self-attention weights for the past 13 days of data on the train set for the one-step-ahead forecast. There is a clear weekly pattern, which peaks on the same day (as the present day) in the previous week (time index -7). Attention increases again as the past day gets closer to the present day (time index 0). Implying the most recent day in the past (time index -1) also has a significant impact on the prediction. 
Figure \ref{fig:attention_holiday} presents the average daily self-attention weights on the train set for the one-step-ahead forecast, along with COVID-19 cases and annotated holidays. We choose the following federal holidays within this highlighted period: (1) Thanksgiving - Nov 26, (2) Christmas Eve - Dec 25, (3) New Year's Day - Jan 1.  It shows that the reported cases have a weekly pattern, which often peaks on Friday and bottoms on Sunday. Similarly, during the holidays, there is a drop in reported cases. The TFT learns such seasonality patterns from the data and the attention weights also show a similar trend where it bottoms on Friday, right before the start of the weekend, and also around the holidays. Then it peaks around Wednesday as we can see a rise in reported cases on Thursday and Friday. 

\subsection{Input Variable Importance}  
We analyze the variable importance by summing up the weights from the Variable Selection Network (VSN) for each variable across the train set. The weights for each feature type are normalized to percentage and presented in Table \ref{table:feature_importance}. For static covariates, the highest weight is given to age distribution, we know the elder population was severely affected by COVID-19. In the observed inputs, past target values (cases, deaths) are the most important as expected. In known future inputs, the county identifier (linear Space) and day of the week (sinweekly, cosweekly) have balanced importance as they all help to learn persistent spatial and temporal patterns.

\begin{table}[!ht]
\centering
\caption{ Feature importance from variable selection weights. The highest values are highlighted in bold.}
\begin{tabular}{l|ccc}
\hline
\textbf{Feature} & \textbf{Static} & \textbf{Observed} & \textbf{Known} \\ \hline
\textbf{Cases} & & \textbf{34.47\%} & \\
Deaths & & 15.94\%& \\ \hline
\textbf{Age Distribution} & \textbf{62.04\%} &   & \\
Health Disparities  & 37.96\% &   & \\ \hline
Vaccination& &  7.90\% &  \\ 
Disease Spread & & 6.22\% & \\
Transmissible Cases & & 4.49\% & \\
Social Distancing & & 4.85\% & \\ \hline
SinWeekly & & 13.83\% & 37.04\% \\
CosWeekly & & 8.66\% & 23.36\%\\
\textbf{Linear Space} & & 3.65\% & \textbf{39.60\%}\\
\hline
\end{tabular}
\label{table:feature_importance}
\end{table}

\section{Feature Sensitivity Analysis}
\label{sec:feature_sensitivity}
\subsection{Model Sensitivity of Individual Feature}
\label{subsec:individualMorris}
Our TFT model is trained using a variety of features, shown in Table \ref{table:features}. Intuitively, these features should have different impacts on the infection of COVID-19. To detect such influence, we studied the sensitivity of individual features using the Morris method. The normalized Morris index $\hat{\mu^{*}}$ of each feature is computed which is introduced in Section III. Since the counties differ from each other drastically in the feature space, to measure feature importance in a more accurate way, we scale $\hat{\mu^{*}}$ by the standard deviation of each input feature $\sigma_i$, and $\hat{\mu^{*}} * \sigma_i$ is used to measure the sensitivity of input features on each county at a single timestamp. We refer to $\hat{\mu^{*}} * \sigma_i$ as \textbf{Scaled Morris Index} in the following texts. Features with a higher scaled Morris index have a greater impact on the output.


\begin{figure}[!ht]
    \centering
    \includegraphics[width=0.40\textwidth]{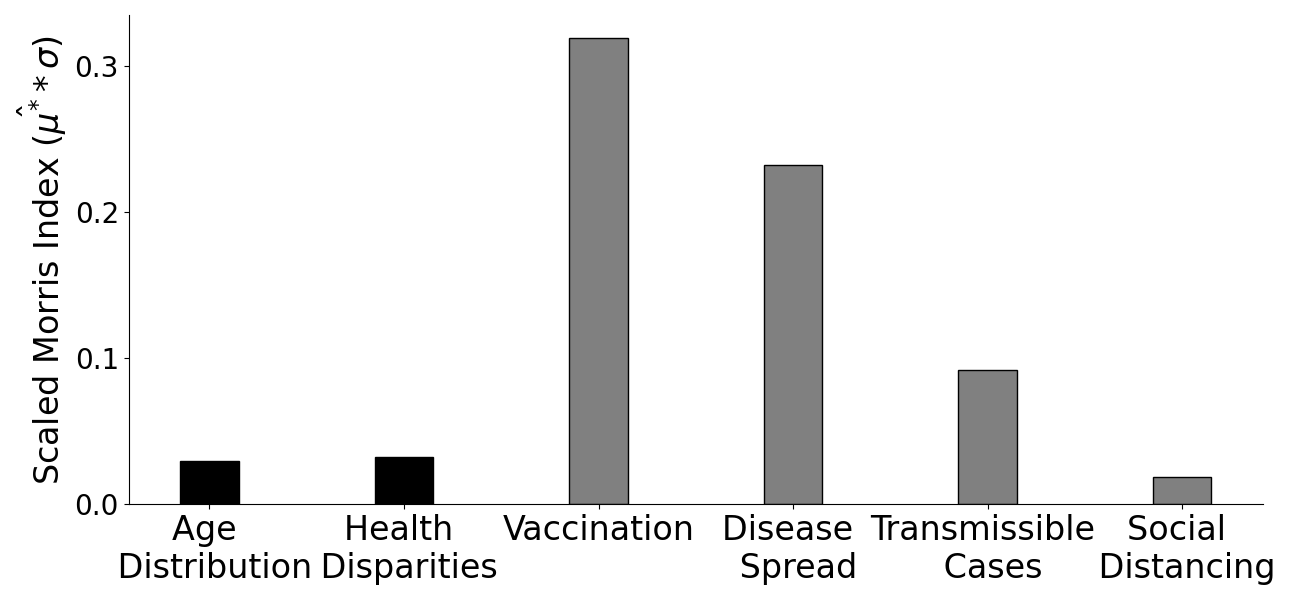}
    \caption{Sensitivity of observed features}
    \label{fig:sensitivity}
\end{figure}
Figure \ref{fig:sensitivity} is for our sensitivity analysis experiments.  Vaccination and disease spread are the most influential features, while other features have much less influence on the output. This result gives proof of the impact of vaccination on COVID-19.

\subsection{Sensitivity of Vaccination to Cases by Feature Subgroups}
In the following two subsections, we discuss in detail our Morris experiments and the results on population in age subgroups and population in industry sectors as static features respectively. While both vaccination and disease spread are influential dynamic features, we choose to focus on studying the impact of vaccination on finer-grained population groups. In this study, we experimented with 2 population segmentation methods: 1) Population segmentation by age, and 2) Population segmentation by industry sectors. 


\begin{figure*}[!h]
    \centering
    \begin{minipage}{0.5\textwidth}
        \centering
        \tabcaption{County level age statistics \cite{pop:2020}.}
        \fontsize{8}{7}\selectfont
        \begin{tabular}{l|l|l|l} 
        \hline
        \textbf{Age} & \textbf{Mean}  & \textbf{Std} & \textbf{Ratio} \\ \hline
        0-19 & 0.242 &  0.036 & 6.622 \\
        20-29 & 0.121 & 0.030 & 3.997 \\
        30-39 & 0.118 & 0.017 & 6.854 \\
        40-49 & 0.114 & 0.012 & 8.886 \\
        50-64 & 0.200 & 0.023 & 8.633 \\
        65-79 & 0.153 & 0.036 & 4.216 \\
        80+ & 0.049 & 0.015 & 3.145 \\
        \hline
        \end{tabular}
        \label{table:age_statistics}
    \end{minipage}
    \begin{minipage}[h]{0.35\textwidth}
    \centering
    \includegraphics[width=\textwidth]{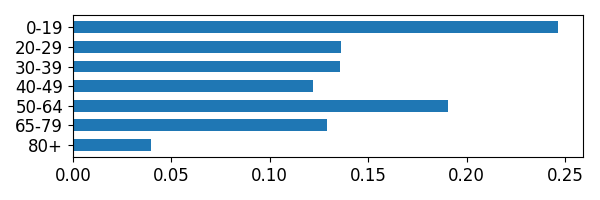}
    \figcaption{Percent of age group in US.}
    \label{fig:Age_Hist}
    \end{minipage}
\end{figure*}

\begin{figure*}[!h]
    \centering
    \begin{minipage}[h]{\textwidth}
    \centering
        \subfloat[Age]{\includegraphics[width=0.35\textwidth]{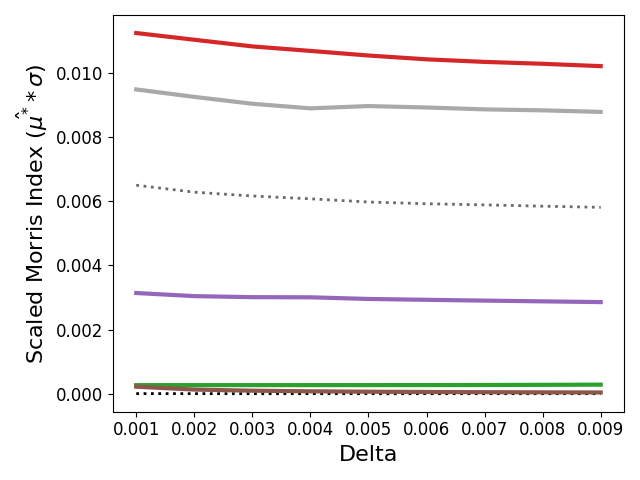}\label{fig:agesens_age}}
        \hspace{5ex}
        \subfloat[Vaccination]{\includegraphics[width=0.35\textwidth]{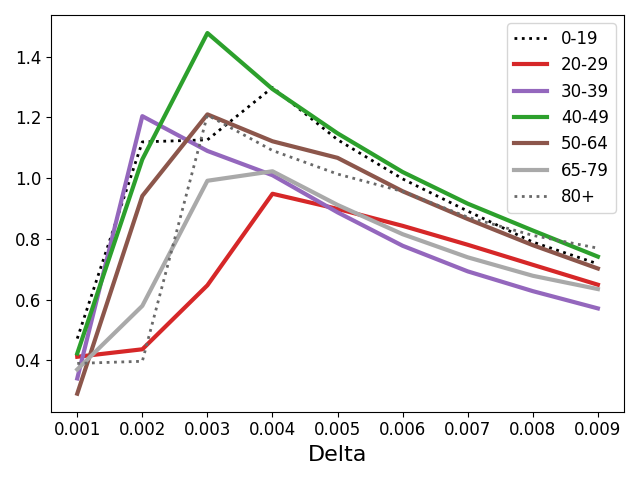}\label{fig:agesens_vaccination}}
        \figcaption{Age subgroup sensitivity results (the greater the value, the more important the feature sensitivity). We study the features that measure a fraction of the county population by age. We then plot the scaled Morris index that multiplies the derivative with the standard deviation but only for those features with larger means. }
        \label{fig:sensitivity_age_subgroup}
    \end{minipage}
\end{figure*}

A total of 7 age subgroups and 11 industry sectors were evaluated in our experiments. We trained our model for each of the population subgroups as the static input and vaccination as the dynamic input, forming a set of 18 individually trained models using TFT with the same experimental setup as described in section \ref{sec:exp_setup}. Table \ref{table:morris_experiments} lists the feature combinations and the training losses of a subset of the trained models.
\begin{table}[!ht]
\centering
\caption{Sensitivity analysis models and training loss}
\begin{adjustbox}{max width=\textwidth}
\begin{tabular}{l|l|l}
\hline
\textbf{Feature Subgroup} & \textbf{Training Loss} & \textbf{$\hat{\mu^{*}} * \sigma$ ($\Delta=0.005$)} \\ \hline
0-19  & 1.50E-04 & 7.20E-07 \\
20-29 & 1.50E-04 & 1.05E-02 \\
30-39 & 1.50E-04 & 2.95E-03 \\
40-49 & 1.49E-04 & 2.68E-04 \\
50-64 & 1.51E-04 & 6.22E-05 \\
65-79 & 1.50E-04 & 8.96E-03 \\
80+ & 1.52E-04 & 5.97E-04 \\
\hline
\hline
Healthcare & 1.49E-04 & 3.53E-03 \\
Manufacturing & 1.61E-04 & 7.92E-03 \\
Retail & 1.50E-04 & 3.26E-03 \\
Accommodation & 1.51E-04 & 5.43E-03 \\
\hline
\end{tabular}
\end{adjustbox}
\label{table:morris_experiments}
\end{table}

\subsection{Sensitivity of Vaccination by Age Population Groups}
\label{sec:sens_age}
CDC evaluated the association between county-level COVID-19 vaccine uptake and rates of COVID-19 cases and deaths in the US from January 1, 2020, to October 31, 2021\cite{mclaughlin2022county}. The results showed US counties with a percent of persons with $\ge$ 80\% of their residents $\ge$12 years of age fully vaccinated against COVID-19 had 30\% and 46\% lower rates of COVID-19 cases and death respectively. Our sensitivity study in Fig. \ref{fig:sensitivity} also identifies vaccination as the most influential feature. In accordance with CDC's recent evaluation \cite{mclaughlin2022county}, we divided the population into 7 age groups reported in Table \ref{table:age_statistics}: 0-19, 20-29, 30-39, 40-49, 50-64, 65-79, 80+. The percentage of the population in each age group across all US counties is reported in Fig. \ref{fig:Age_Hist}.

\begin{figure*}[!h]
    \centering
    \begin{minipage}{0.45\textwidth}
        \centering
        \tabcaption{County level industry statistics\cite{cbp2020}.}
        \begin{adjustbox}{max width=\textwidth}
        \begin{tabular}{l|l|l|l} 
        \hline
        \textbf{Industry} & \textbf{Mean}  & \textbf{Std} & \textbf{Ratio} \\ \hline
        \textbf{Health Care and Social Assistance} & \textbf{0.18} & \textbf{0.09} & 0.5 \\
        \textbf{Manufacturing} & \textbf{0.15} & \textbf{0.13} & 0.85 \\
        \textbf{Retail Trade} & \textbf{0.16} & \textbf{0.07} & 0.42 \\
        \textbf{Accommodation and Food Services} & \textbf{0.12} & \textbf{0.07} & 0.62 \\
        Mining, Quarrying, and Oil and Gas Extraction & 0.02 & 0.06 & 3.6 \\
        Construction & 0.06 & 0.05 & 0.87 \\
        Transportation and Warehousing & 0.04 & 0.05 & 1.08 \\
        Professional, Scientific, and Technical Services & 0.04 & 0.04 & 1.05 \\
        Wholesale Trade & 0.04 & 0.05 & 1.08 \\
        Administrative and Support & 0.04 & 0.05 & 1.29 \\
        Educational Services & 0.01 & 0.03 & 2.16 \\
        Utilities & 0.01 & 0.03 & 3.35 \\
        Finance and Insurance & 0.04 & 0.03 & 0.74 \\
        Agriculture, Forestry, Fishing and Hunting & 0.01 & 0.02 & 3.56 \\
        Arts, Entertainment, and Recreation & 0.01 & 0.03 & 1.97 \\
        Management of Companies and Enterprises & 0.01 & 0.02 & 2.47 \\
        Other Services (except Public Administration) & 0.05 & 0.03 & 0.59 \\
        Information & 0.01 & 0.02 & 1.24 \\
        \hline
        \end{tabular}
        \end{adjustbox}
        \label{table:statindustry}
    \end{minipage}
    \begin{minipage}[h]{0.45\textwidth}
    \centering
    \includegraphics[width=\textwidth]{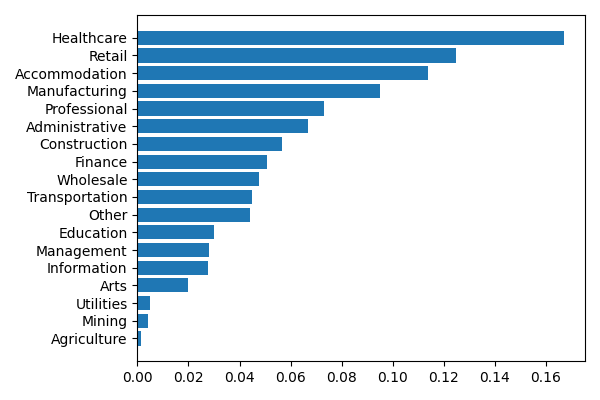}
    \figcaption{Percent of population in industry sectors.}
    \label{fig:Industry_Hist}
    \end{minipage}
\end{figure*}

\begin{figure*}[!h]
    \centering
    \begin{minipage}[h]{\textwidth}
    \centering
    \subfloat[Industry]{\includegraphics[width=0.35\linewidth]{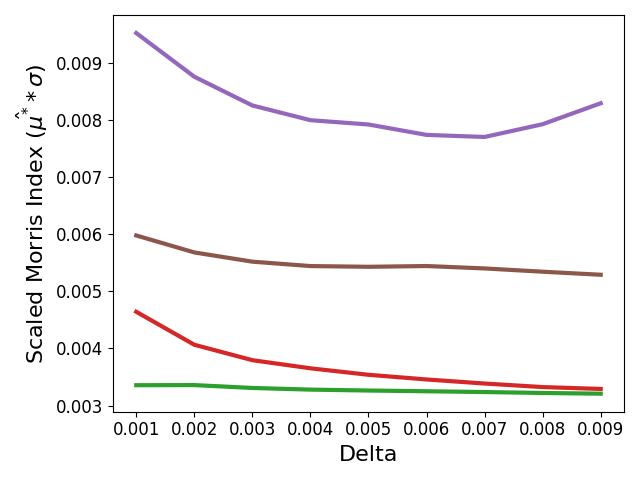}\label{fig:industrysens_industry}}
    \hspace{5ex}
    \subfloat[Vaccination]{\includegraphics[width=0.35\linewidth]{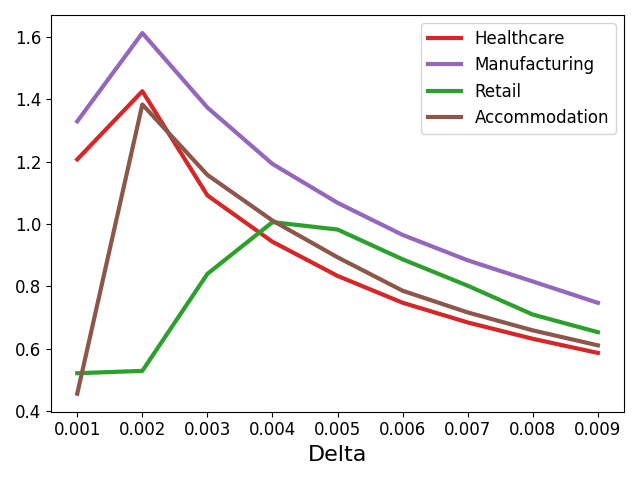}\label{fig:industrysens_vaccination}}
    \caption{Industry sectors and sensitivity results (the greater the value, the more important the feature sensitivity). We study features that measure a fraction of the county population by industry groups. We favor those with a large mean as they represent a larger proportion of the observed data. Quantitatively, we plot the scaled Morris index that multiplies the derivative with the standard deviation but only for those features with larger means (the top 4 industry sectors). Manufacture shows the most impact, followed by Accommodations in the Industry plot (left). The Vaccination plot (right) is less distinguishable between the sectors due to convergence at a large delta in the Morris indices. }
    \label{figure:sensitivity_industry_subgroup}
    \end{minipage}
\end{figure*}

Fig. \ref{fig:sensitivity_age_subgroup} shows Morris experiments and results on the age subgroups. The solid lines specify working age groups and dotted lines represent the non-working age groups. Figure \ref{fig:agesens_age} shows the sensitivity of age groups to COVID-19 cases. Age groups 20-29, 65-79, and 80+ have larger Morris indices than other age groups, which indicates that the population from these groups is more sensitive to COVID-19 cases as changes in their population lead to greater changes in COVID-19 cases.
The sensitivity of vaccination to COVID-19 cases is shown in Figure \ref{fig:agesens_vaccination}. Vaccination is less distinguishable because all age subgroups converge at Morris indices at large delta values. The 40-49 and 0-19 age groups have slightly more impact on the cases than the 20-29 and 65-79 age groups in our experiments. This indicates that increasing the vaccination rate among these population groups will be equally impactful in COVID-19 cases.

\subsection{Sensitivity of Vaccination by Industry Sectors}
\label{sec:sens_industry}
We also studied the sensitivity of vaccination in our model by segmenting the population into the 18 majority represented industry sectors in Table \ref{table:statindustry}. To further rank the county-level population, Figure \ref{fig:Industry_Hist} represents the primary workplace setting across the 3142 US counties. The secondary Industry segments provide further details on common jobs in each of the 18 major segments \cite{industry2017}, based on North American Industry Classification System (NAICS) from the Us Census Bureau. Note that the "Other Services" segment covers the rest of the industry jobs that cannot fit on the Table. Industry sectors in bold texts in Table \ref{table:statindustry} are studied in our experiments to indicate a higher population segment. The industry subgroups enable us to better understand the impact of COVID-19 in workplace settings. Looking at  Fig. \ref{fig:industrysens_industry}, among the 4 industry sectors, increasing the Morris index in the Manufacturing sector seems to be the most sensitive to the infection, followed by Accommodation, Healthcare, and Retail. 

\section{Related Work \label{sec:related_works}}

\subsection{Statistical and Machine Learning Models} 
Many efforts have been made into COVID-19 forecasting using statistical learning, epidemiological, and deep learning models \cite{ wang2020time, chimmula2020time,zeroual2020deep }. Different statistical and mathematical models such as Susceptible Infectious Recovery (SIR) and Susceptible Exposed Infectious Recover (SEIR), have been used to simulate the COVID-19 spread \cite{chimmula2020time, weitz2020modeling}. These models provide useful insights, however, their performance is limited by the number of complex influencing factors and relationships they can capture, which is where the machine learning based models excel. Despite the difficulties in interpretability, deep learning models generally produce a better performance \cite{shahid2020predictions,wang2020time} than the traditional machine learning and time series forecasting models, such as ARIMA\cite{kirbacs2020comparative}, SVR\cite{shahid2020predictions}, and XGBoost\cite{luo2021time}.

\subsection{Deep Learning Models and COVID-19 Forecasting} 
Deep learning has shown significant performance in time series forecasting \cite{lim2021temporal} and has been widely used to predict COVID-19 infection. LSTM and Bi-LSTM based models often outperform other approaches in time-series forecasting \cite{shahid2020predictions, chimmula2020time,kirbacs2020comparative} because RNN-based models are more suitable for time series data with Spatio-temporal sequences \cite{chandra2022deep}. Furthermore, \cite{zeroual2020deep} found that Variational Auto Encoder (VAE) outperforms several other deep learning models in predicting daily confirmed and recovered COVID-19 cases. AICov \cite{fox_von_laszewski_wang_pyne_2021} and \cite{hssayeni2021forecast} used LSTM to predict COVID-19 in US counties. New research using the AI-based predictions also discusses 1) new virus variants are being discovered\cite{rashed2021infectivity}, 2) data quality and quantity used for model training are limited\cite{weitz2020modeling,saez2021potential}, 3) ML-based models are not guaranteed to take socioeconomic, cultural and demographic factors into consideration while learning the data\cite{saez2021potential}.

\subsection{Interpreting COVID-19 Forecasting Models} 
With the increasing use of deep learning models, their interpretation has gained increasing demand \cite{lim2021time} to understand the model's decisions. DeepCOVID framework \cite{rodriguez2021deepcovid} used RNN with auto-regressive inputs to predict COVID-19 cases and interpret the contribution of input signals in prediction performance. DeepCOVIDNet \cite{ramchandani2020deepcovidnet} analyzed the features and their interactions in predicting the range of infected cases increases. Self-Adaptive Forecasting \cite{arik2022self} is a novel method to adapt models on non-stationary time series data as well as giving interpretations with TFT. 

\section{Conclusions and Future Work \label{sec:conclusion}} 
This paper uses deep learning to address the challenging problem of model feature sensitivity, which is critical in improving the interpretation of forecasting. With the COVID-19 data, we find that TFT performs well in learning temporal patterns from the data. The ”self-attention” architecture sets up a strong global dependency structure. While it learns long-range dependencies, it is less effective in highly dynamic forecasting problems with non-stationary sequences, such as extreme or unknown events. However, the TFT model provides an excellent test environment enabling us to look into the COVID-19 feature sensitivity of high volatile time series and static variables. Combined with the Morris method, this sensitivity study enables us to look into stratifying the modeling based on population subgroups such as industry and age and see how that affects  COVID-19 responses to cases of infection and other important features at the community level. 

The conceptualization of modeling individual features with the post hoc sensitivity analysis can apply to other time series models as the Morris method is efficient and generic. The advantages of using fewer features have enabled one to get insights into refined strategy optimizations. The Morris index and our novel individual multidimensional feature importance evaluation in this paper can contribute to autonomous feature engineering in forecasting and other similar challenges. COVID-19 infection prediction is a critical tool for policymakers responding to a global health challenge with observed data changing in real-time. Future works can include analyzing the COVID-19 feature importance and descriptive accuracy in many health and financial problems. 

\section*{Acknowledgement}
This  work is supported by NSF grant CCF-1918626  Expeditions: Collaborative  Research: Global Pervasive Computational Epidemiology, and NSF Grant 1835631 for CINES: A Scalable Cyberinfrastructure for Sustained Innovation in Network Engineering and Science.

\bibliographystyle{IEEEtran}
\bibliography{IEEEabrv,bibliography}

\end{document}